\makeatother
\pdfoutput=1
\documentclass[sigconf, nonacm]{acmart}
\usepackage{balance}
\usepackage{graphicx}
\usepackage{amsfonts}

\usepackage{amssymb}
\usepackage{amsmath}
\usepackage{bm}
\usepackage{multirow}
\usepackage{booktabs}
\usepackage{ulem}
\usepackage{float}
\usepackage{makecell}
\usepackage{colortbl}
\usepackage{xcolor}
\usepackage{algorithm}
\usepackage{algpseudocode}

\usepackage{xspace}
\makeatletter
\renewcommand{\maketag@@@}[1]{\hbox{\m@th\normalsize\normalfont#1}}%
\algnewcommand\algorithmicforeach{\textbf{for each}}
\algdef{S}[FOR]{ForEach}[1]{\algorithmicforeach\ #1\ \algorithmicdo}
\newcommand\vldbdoi{XX.XX/XXX.XX}

\usepackage{pifont}

\newcommand{\method}{\textsc{GCMAE}\xspace}

\newcommand{\stitle}[1]{\vspace*{0.4em}\noindent{\bf #1\/}}

\begin{document}
\title{Generative and Contrastive Paradigms Are Complementary for \\Graph Self-Supervised  Learning}


\author{Yuxiang Wang$^{\dagger}$ \quad Xiao Yan$^{\#}$ \quad Chuang Hu$^{\dagger}$ \quad Fangcheng Fu$^{\ddagger}$ \quad Wentao Zhang$^{\ddagger}$ \\
Hao Wang$^{\dagger}$ \quad Shuo Shang$^{\S}$  \quad Jiawei Jiang$^{\dagger}$}
\affiliation{%
  \institution{
  $^{\dagger}$ School of Computer Science, Wuhan University \quad $^{\#}$ Southern University of Science and Technology \\
  $^{\ddagger}$ Peking University \quad $^{\S}$ University of Electronic Science and Technology of China
  }
}
\email{{nai.yxwang,handc,wanghao.cs,jiawei.jiang}@whu.edu.cn}
\email{yanxiaosunny@gmail.com}
\email{{ccchengff,wentao.zhang}@pku.edu.cn}  \email{jedi.shang@gmail.com}


\begin{abstract}
For graph self-supervised learning (GSSL), \textit{masked autoencoder} (MAE) follows the generative paradigm and learns to reconstruct masked graph edges or node features. \textit{Contrastive Learning} (CL) maximizes the similarity between augmented views of the same graph and is widely used for GSSL. However, MAE and CL are considered separately in existing works for GSSL. We observe that the MAE and CL paradigms are complementary and propose the \textit{graph contrastive masked autoencoder} (\method) framework to unify them. Specifically, by focusing on local edges or node features, MAE cannot capture global information of the graph and is sensitive to particular edges and features. On the contrary, CL excels in extracting global information because it considers the relation between graphs. As such, we equip \method~ with an MAE branch and a CL branch, and the two branches share a common encoder, which allows the MAE branch to exploit the global information extracted by the CL branch. To force \method~ to capture global graph structures, we train it to reconstruct the entire adjacency matrix instead of only the masked edges as in existing works. Moreover, a discrimination loss is proposed for feature reconstruction, which improves the disparity between node embeddings rather than reducing the reconstruction error to tackle the feature smoothing problem of MAE. We evaluate \method~ on four popular graph tasks (i.e., node classification, node clustering, link prediction, and graph classification) and compare it with 14 state-of-the-art baselines. The results show that \method~ consistently provides good accuracy across these tasks, and the maximum accuracy improvement is up to 3.2\% compared with the best-performing baseline.
\end{abstract}

\maketitle



\section{Introduction}

Graphs model entities as nodes and the relations among the entities as edges and prevail in many domains such as social networks~\cite{hamilton2017inductive,velivckovic2017graph}, finance~\cite{liu2021pick}, biology~\cite{dai2018learning}, and medicine~\cite{liu2020hybrid}. By conducting message passing on the edges and utilizing neural networks to aggregate messages on the nodes, graph neural network (GNN) models perform well for various graph tasks, e.g., node classification, node clustering, link prediction, and graph classification~\cite{guo2022graph, zhu2021graph, hamilton2017inductive,zhang2020reliable,zheng2022bytegnn}. 
These tasks facilitate many important applications including recommendation, fraud detection, community detection, and pharmacy~\cite{lu2022bright,gao2021ics,cui2021metro,chen2023ics,chen2023communityaf}. 
To train GNN models, \textit{graph self-supervised learning} (GSSL) is becoming increasingly popular because it does not require label information~\cite{you2020graph,yin2022autogcl}, which can be rare or expensive to obtain in practice. Existing works handle GSSL with two main paradigms~\cite{wu2021self,li2023seegera}, i.e., \textit{masked autoencoder} (MAE) and \textit{contrastive learning} (CL).

Graph MAE methods usually come with an encoder and a decoder, which are both GNN models~\cite{hou2022graphmae,hou2023graphmae2,li2022maskgae,li2023seegera,tan2023s2gae}.
The encoder computes node embeddings from a corrupted view of the graph, where some of the edges or node features are masked; the decoder is trained to reconstruct the masked edges or node features using these node embeddings. Graph MAE methods achieve high accuracy for graph tasks but we observe that they still have two 
limitations. \ding{182} Graph MAE \textit{misses global information} of the graph, which results in sub-optimal performance for tasks that require a view of the entire graph (e.g., graph classification)~\cite{chien2021node,wei2022masked}. This is because both the encoder and decoder GNN models use a small number of layers (e.g., 2 or 3), and a \textit{k}-layer GNN can only consider the \textit{k}-hop neighbors of each node~\cite{chen2020measuring}. Thus, both the encoder and decoder consider the local neighborhood. \ding{183} Graph MAE is prone to \textit{feature smoothing}, which means that neighboring nodes tend to have similar reconstructed features and harms performance for tasks that rely on node features (e.g., node classification). 
This is because the encoder and decoder GNN models aggregate the neighbors to compute the embedding for each node, and due to this local aggregation, GNN models are widely known to suffer from the over-smoothing problem~\cite{zhao2019pairnorm,chen2020measuring}, which also explains why GNN cannot use many layers.

Graph CL methods usually generate multiple perturbed views of a graph via data augmentation and train the model to maximize the similarity between positive pairs (e.g., an anchor node and its corresponding node in another view) and minimize the similarity between negative pairs (e.g., all other nodes except the positive pairs)~\cite{li2022let,cai2023lightgcl}.
As graph CL can contrast nodes that are far apart (e.g., > 3 hops) in multiple views, it captures global information of the entire graph. However, without reconstruction loss for the node features and edges, CL is inferior to MAE in learning local information for the nodes or edges, and thus may have sub-optimal performance for tasks that require such information (e.g., link prediction and clustering)~\cite{you2020graph,you2021graph}.

\begin{figure}[t]
    \centering
    \includegraphics[width=\linewidth]{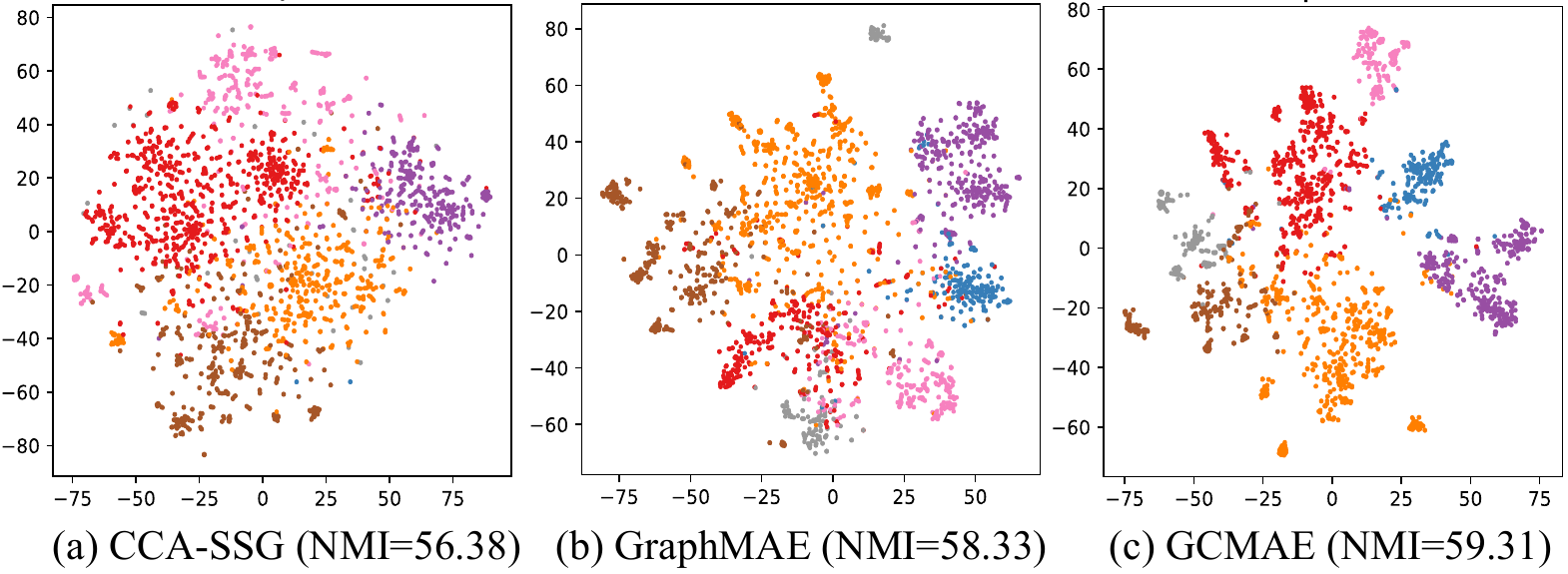}
    \caption{We conduct node clustering on the Cora and visualize the node embeddings learned by our GCMAE, GraphMAE, and CCA-SSG with t-SNE. GraphMAE and CCA-SSG represent MAE and CL methods, respectively. Nodes of the same category are plotted in the same color. We use Normalized Mutual Information (NMI) to evaluate the clustering effect, with a larger value indicating better accuracy.}
    \label{fig:clustering}
\end{figure}

The above analysis shows that graph MAE and CL methods potentially complement each other in capturing local and global information of the graph. Thus, combining MAE and CL may yield better performance than using them individually. Figure~\ref{fig:clustering} shows such an example with node clustering. As a representative graph CL method, CCA-SSG~\cite{zhang2021canonical} has poor clustering for the nodes due to the lack of the local information. For GraphMAE~\cite{hou2022graphmae}, a state-of-the-art graph MAE method, the node clusters are more noticeable by capturing local information. However, by combining MAE and CL, our \method~ yields the best node clustering and hence the highest accuracy among the three methods. Although the idea sounds straightforward, a framework that combines MAE and CL needs to tackle two key technical challenges. \ding{182} MAE and CL have different graph augmentation logic and learning objectives (i.e., reconstruction and discrimination), and thus a unified view is required to combine them. \ding{183} MAE and CL focus on local and global information respectively, so the model architecture should ensure that the two kinds of information complement each to reap the benefits.  
\begin{table}[t]
\centering
\caption{Performance improvements of our \method~over the best-performing baselines in each category for different tasks.
\textit{Others} are specialized methods for each task.
}
\label{tab:improvent across tasks}
\begin{tabular}{c|ccc}
\toprule
  Graph Task                   & vs. Contrastive & \quad vs. MAE \quad  & \quad Others \quad \\
\midrule
Node classification  & 4.8\%      & 2.2\% & 12.0\% \\
Link prediction      & 4.4\%      & 1.5\% & -\\
Node clustering      & 8.8\%      & 3.2\% & 14.7\% \\
Graph classification & 2.5\%      & 4.2\% &-\\
\bottomrule
\end{tabular}
\end{table}

To tackle these challenges, we design the \textit{graph contrastive masked autoencoder} (i.e., \method~) 
framework. We first show that MAE and CL can be unified in one mathematical framework despite their apparent differences. In particular, we express MAE as a special kind of CL, which uses masking for data augmentation and maximizes the similarity between the original and masked graphs. This analysis inspires us to use \textit{an MAE branch} and \textit{a CL branch} in \method~ and share \textit{the encoder} for the two branches. 
The MAE branch is trained to reconstruct the original graph while the CL branch is trained to contrast two augmented views; and the shared encoder is responsible for mapping the graphs (both original and augmented) to the embedding spaces to serve as inputs for the two branches. By design, the shared encoder conducts input mapping and transfers information between both branches. As such, local information and global information are condensed in the shared encoder and benefit both MAE and CL. Furthermore, to tackle the feature smoothing problem of MAE that makes the reconstructed features of neighboring nodes similar,
we introduce a \textit{discrimination loss} for model training, which enlarges the variance of the node embeddings to avoid smoothing. 
Instead of training MAE to reconstruct only the masked edges as in existing works, we reconstruct the entire adjacency matrix such that MAE can learn to capture global information of the graph structures.

We evaluate our \method~on four popular graph tasks, i.e., node classification, node clustering, link prediction, and graph classification.
These tasks have essentially different properties and require different information in the graph.
We compare \method~ with state-of-the-art baselines, including graph MAE methods such as GraphMAE~\cite{hou2022graphmae}, SeeGera~\cite{li2023seegera}, S2GAE~\cite{tan2023s2gae}, and MaskGAE~\cite{li2022maskgae}, and graph CL methods such as GDI~\cite{velivckovic2017graph}, MVGRL~\cite{hassani2020contrastive}, GRACE~\cite{zhu2020deep}, and CCA-SSG~\cite{zhang2021canonical}. The results show that \method~yields consistently good performance on the four graph tasks and is the most accurate method in almost all cases (i.e., a case means a dataset plus a task). As shown in \autoref{tab:improvent across tasks}, \method~outperforms both contrastive and MAE methods, and the improvements over the best-performing baselines are significant. 
We also conduct an ablation study for our model designs (e.g., shared encoder and the loss terms), and the results suggest that the designs are effective in improving accuracy. 

To summarize, we make the following contributions:
\begin{itemize}
\item
We observe the limitations of existing graph MAE and CL methods for graph self-supervised learning (GSSL) and propose to combine MAE and CL for enhanced performance.

\item We design the \method~framework, which is the first to jointly utilize MAE and CL for GSSL to our knowledge.

\item  We equip \method~ with tailored model designs, e.g., the \textit{shared encoder} for information sharing between MAE and CL, the \textit{discrimination  loss} to combat feature smoothing, and training supervision with \textit{adjacency matrix reconstruction}.

\item We conduct extensive experiments to evaluate \method~ and compare it with state-of-the-art, demonstrating that \method~ is general across graph tasks and effective in accuracy.
\end{itemize}

\begin{figure*}[t]
    \centering
    \includegraphics[width=\linewidth]{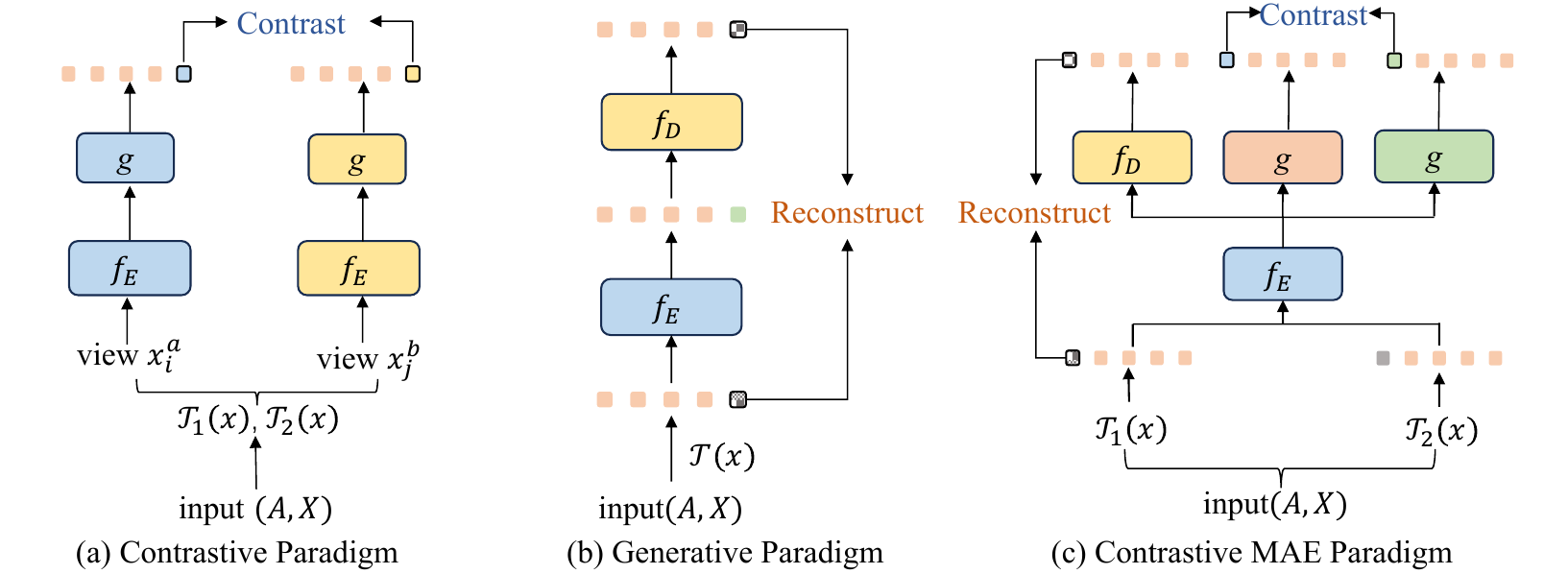}
    \caption{Architectures of  the generative paradigm, contrastive paradigm, and our \method~ for graph self-supervised learning. \textit{Contrast} means the contrastive loss while \textit{Reconstruct} means the  reconstruction loss.}
    \label{fig:SSL}
\end{figure*}


\section{PRELIMINARIES}
In this part, we introduce the basics of MAE and CL for GSSL to facilitate our formulation. The discussions on specific MAE and CL methods are provided in Section~\ref{sec:related}. We use $G=(\bm{V}, \bm{E})$ to denote a graph, where $\bm{V}=\{v_1,v_2,\cdot \cdot \cdot,v_N\}$ is the set of nodes with $|\bm{V}|=N$ and $\bm{E}$ is the set of edges.
The adjacency matrix of the graph is $\bm{A}\in \{0,1\}^{N\times N}$, and the node feature matrix is $\bm{X}\in \mathbb{R}^{N\times d} $, where $x_i\in\mathbb{R}^d $ is the feature vector of $v_i$ with dimension $d$ and $\bm{A}_{ij}=1$ if edge $(v_i,v_j)\in \bm{E}$.


\begin{figure*}[t]
    \centering
\includegraphics[width=\linewidth,scale=1.00]{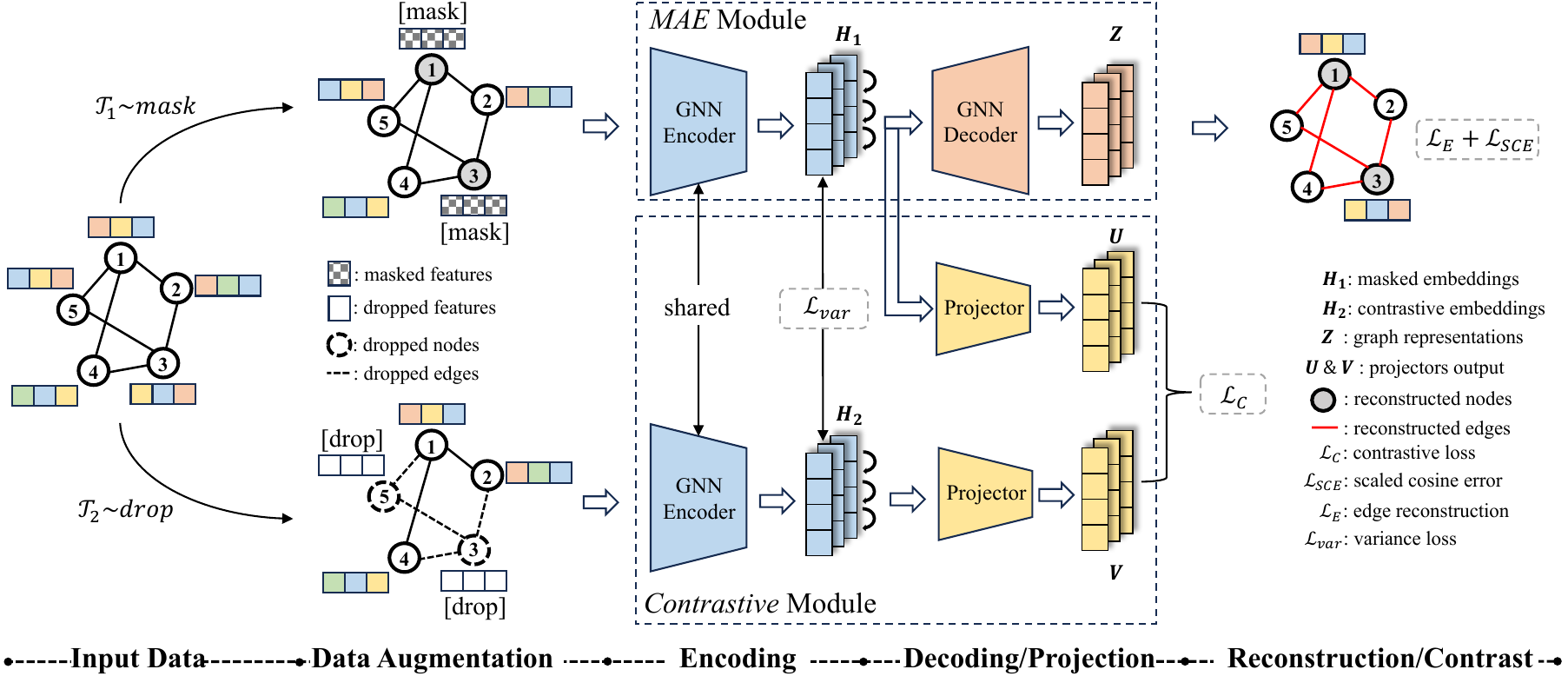}
    \caption{Structure of our \method~ framework, which consists of an MAE branch and a contrastive branch with a shared encoder.}
    \label{fig:framework}
\end{figure*}

\subsection{Contrastive Learning (CL) for GSSL}
CL has been very popular for GSSL due to its good performance for various application scenarios, such as social network analysis~\cite{zhang2021canonical}, molecular detection~\cite{ma2021contrastive}, and financial deception~\cite{sahu2020ubiquity}.
Generally, graph CL follows a ``augmenting-contrasting'' pattern,
where the graph views are corrupted via the data augmentation strategies $\mathcal{T}_1$ and $\mathcal{T}_2$ (e.g., node feature masking and node dropping), and then a GNN encoder $f_{\theta}$ is used to encode the corrupted views into node embeddings.
The goal of CL is to maximize the mutual information between the two corrupted views, which is achieved by maximizing the similarity of node embeddings as a surrogate.
Thus, CL does not require label information and conducts learning by distinguishing between node embeddings in different views.
\autoref{fig:SSL} (a) shows the framework of graph contrastive learning.
The objective of CL is essentially to maximize the \textit{similarity function} between two node embeddings encoded through a GNN encoder $f_{\theta}$, which can be formulated as: 
\begin{equation}
        \max_{\theta } \mathop{\mathbb{E}} \limits_{x\sim \mathcal{D} }S\left ( f_{\theta} \left ( \mathcal{T}_1\left (\bm{A}, \bm{X} \right )   \right ),f_{\theta} \left ( \mathcal{T}_2\left ( \bm{A}, \bm{X} \right )   \right ) \right ) ,
        \label{eq:con paradigm}
\end{equation}
where $\theta$ represents the network weights, node feature vector $x$ follows the input data distribution $\mathcal{D}$, and $S\left ( \cdot ,\cdot  \right ) $ is the similarity function to measure the similarity between node embeddings.

\subsection{Masked Autoencoder (MAE) for GSSL}

Different from graph CL methods, graph MAE follows a ``masking-reconstructing'' pattern.
The overview of graph MAE methods is shown in \autoref{fig:SSL} (b).
In general, graph MAE methods first randomly mask node features or edges with mask patches $\bm{M}$, where $x\odot \bm{M}$ represents visible tokens and $x\odot (1-\bm{M})$ means masked tokens.
Graph MAE leverages the encoder $f_\theta$ to encode the visible tokens into node embeddings, and then the decoder $d_\phi$ attempts to reconstruct the masked tokens by decoding from the node embeddings.
Thus, we wish to maximize the similarity between the masked tokens and the reconstructed tokens:
\begin{equation}
    \max_{\theta,\phi  } \mathop{\mathbb{E}} \limits_{x\sim \mathcal{D} }S\left ( d_\phi \left ( h \right )\odot (1-\bm{M}) ,x\odot (1-\bm{M}) \right ) , h=f_\theta \left ( x\odot \bm{M}  \right ),
    \label{eq:mae paradigm}
\end{equation}
where $\odot$ means element-wise product, $h$ is the node embedding learned by the encoder, and $S\left (\cdot ,\cdot   \right ) $ is the \textit{similarity function} for MAE modeling.

\section{The GCMAE Framework}

Since graph MAE cannot benefit from global information, it can only learn from limited neighbor nodes, which may eventually lead to similar graph representations.
We are inspired by the successful application of CL, where global information can be learned by contrasting the anchor node with distant nodes.
The overall framework is shown in \autoref{fig:framework}.

\subsection{Unifying CL and MAE for GSSL}
\label{sec:theory}

Even though contrastive and generative approaches have achieved individual success, there is a lack of systematic analysis regarding their correlation and compatibility in one single framework.
Motivated as such, we aim to explore a \textit{unified} Contrastive MAE paradigm to combine contrastive paradigm and MAE paradigm.
From \autoref{eq:mae paradigm}, we can conclude that graph MAE essentially maximizes the similarity between node embeddings reconstructed via decoder and the masked tokens, which is formally analogous to \autoref{eq:con paradigm}.
In other words, both the contrastive paradigm and the generative paradigm are maximizing the similarity between two elements within the function.

Let us still focus on \autoref{eq:mae paradigm}, graph MAE wishes to find a theoretically optimal decoder that can reconstruct the masked tokens losslessly.
Suppose we can achieve the decoder $d^{}{}'_{\phi^{}{}'}$ parameterized by $\phi^{}{}'$ that satisfies $d^{}{}'_{\phi^{}{}'}\left ( f_\theta \left ( x\odot (1-\bm{M})\right )  \right )\cdot(1-\bm{M})\approx x \odot \left ( 1-\bm{M} \right ) $ as closely as possible.
Then, we transform \autoref{eq:mae paradigm} into the following form:
\begin{small}
\begin{equation}
    \max_{\theta, \phi, \phi^{'} }  S( d_\phi( h )  \odot(1\!-\!\bm{M}) \!-\! d^{}{}'_{\phi^{}{}'} ( h)  \odot (1-\bm{M}) ) , 
\label{eq:approximate decoder}
\end{equation}
\end{small}
where $\phi^\prime$ is optimized in the following form:
\begin{small}
    \begin{equation}
        \phi^{}{}'\!=\! {\mathrm {arg}  \max_{\phi^{}{}'}}  \mathop{\mathbb{E}} \limits_{x^{}{}'\sim \mathcal{D} } S( d^{}{}'_{\phi^{}{}'} ( f_\theta ( x^{}{}'\!\odot \!  ( 1\!-\!\bm{M} )))\!\odot\!( 1\!-\!\bm{M}) \!-\!x^{}{}' \!\odot\! ( 1\!-\!\bm{M}  ) ),
    \end{equation}
\end{small}
where $x^{}{}'$ is the feature token reconstructed by the optimal decoder $d^{}{}'_{\phi^{}{}'}$.
Notice that since $d$ and $d^{}{}'$ have the same architecture, we let $d=d^{}{}'$.
Inspired by \cite{kong2023understanding}, we further simplify \autoref{eq:approximate decoder}, and define the loss function for MAE:
\begin{equation}
    \mathcal{L}(h_1,h_2,\phi,\phi^{}{}')= \max_{ \phi, \phi^{'} } S(d_{\phi}(h_1),d_{\phi^{}{}'}(h_2))\odot(1-\bm{M}) ,
    \label{eq: simplified mae}
\end{equation}
where $h_1,h_2$ are the hidden embeddings derived from two augmentation strategies:
\begin{equation}
    \begin{cases}
 & h_1=f_{\theta}(\mathcal{T}_1(x))=f_{\theta}(x\odot \bm{M}), \\ 
        &h_2=f_{\theta}(\mathcal{T}_2(x))=f_{\theta}(x\odot (1-\bm{M})).
\end{cases}
    \label{eq:mae augmentation}
\end{equation}
In this way, we rewrite the generative paradigm into a form similar to the contrastive paradigm, both with data augmentation and optimizing a similarity function.

In order to integrate MAE and CL into one optimization framework, a novel self-supervised paradigm, \textit{Contrastive MAE} paradigm.
The MAE branch is trained to reconstruct the masked tokens using similarity function $S_1$ while the CL branch is trained to contrast two augmented views using similarity function $S_2$:
\begin{equation}
    \mathcal{L}(x,\bm{M},h_1,h_2,\theta,\phi) \!=\! \max_{\theta, \phi }S_{1}(d_{\phi}(h_1),x \!\odot\! (1\!-\!\bm{M})) \!+\!S_{2}(h_1,h_2).
\label{eq:detailed con MAE}
\end{equation}
\noindent \textbf{Summary.}
Despite the recent attempts by contrastive MAE methods~\cite{qi2023contrast,huang2022contrastive}, understanding their underlying rationale remains an open question.
Therefore, we provide a theoretical analysis showing that the nature of the generative paradigm is essentially similar to that of CL.
Rather than treating them separately, these two paradigms can potentially complement each other, compensating for their respective shortcomings and obtaining better performance.
Moreover, both the generative and contrastive paradigms share similar optimization objectives, which enables us to optimize them within a unified framework.

\subsection{Structure of the \method~Framework}
\noindent \textbf{\textit{Objective}}: In this paper, we aim to unify graph MAE and CL into a single framework and further reveal the intrinsic relation between graph MAE and CL, where CL can help graph MAE to achieve global information, through which we hope to further improve the performance of graph MAE on various downstream tasks.

\stitle{\textit{Overview}}: We propose a novel graph SSL framework to incorporate CL into graph MAE.
Overall, our framework consists of two core modules: the MAE Module and the Contrastive Module, which are used to reconstruct masked nodes and contrast two corrupted views, respectively.
To achieve our design goal, we introduce three main components:
\begin{itemize}
    \item \textbf{Shared encoder.} 
    We take the masked graph in Graph MAE as an augmented view, and then generate another view by randomly dropping nodes.
    We leverage a shared encoder to connect both modules and encode the corrupted views into hidden node embeddings.
    In this way, the contrastive module can transfer the global information to the MAE module.
    The node embedding generated from MAE module is used to calculate feature reconstruction loss using Scaled Cosine Error (SCE) $\mathcal{L}_{SCE}$, and another node embedding is leveraged to compute contrastive loss $\mathcal{L}_{C}$.
    \item \textbf{Discrimination loss.}
    Low discriminative input node features will cause graph MAE to generate similar node representations. Thus, we propose a novel discrimination loss $\mathcal{L}_{Var}$, which improves feature discriminability by increasing the variance between node hidden embeddings (i.e., the output of the shared encoder).
    \item \textbf{Adjacency matrix reconstruction.}
    In order to further improve performance on link prediction and node clustering, we reconstruct the entire graph and calculate the adjacency matrix reconstruction error $\mathcal{L}_{E}$.
    This is because reconstructing the limited masked edges is not enough to learn the entire graph structures.
\end{itemize}
Therefore, the overall loss function for GCMAE is defined as follows:
\begin{equation}
    \mathcal{J}=\mathcal{L}_{SCE}+\alpha \mathcal{L}_C +\lambda  \mathcal{L}_E+\mu\mathcal{L}_{Var},
    \label{eq:final loss}
\end{equation}
where $\alpha, \lambda$ and $\mu$ are hyper-parameter used to adjust the weights.
In summary, the contrastive module learns global information by contrasting two corrupted views and transfers it to the MAE module with a shared encoder.
Before decoding, discrimination loss is leveraged to improve the discrimination between node embeddings to avoid yielding similar node representations.
We reconstruct the node features and adjacency matrix in the MAE module.

\section{Model Designs}
In this section, we present a detailed description of the proposed GCMAE.
We introduce each core component individually with the following three questions in our mind:

\begin{itemize}
    \item \textbf{Q1:} How to learn global information for graph MAE?
    
    \item \textbf{Q2:} How to train a decent encoder to learn the entire graph structures?
    
    \item \textbf{Q3:} How to enhance the feature discrimination?
\end{itemize}

Recently, GraphMAE~\cite{hou2022graphmae} has attracted great attention in graph SSL due to its simple but effective framework.
Therefore, we choose GraphMAE as the backbone in this paper. 
For a node set $\bm{V}$, we randomly select a subset of node $\widetilde{\bm{V}} \subseteq  \bm{V}$ and mask their corresponding features $\bm{X}$, i.e., setting its feature values to 0.
We sample $\widetilde{\bm{V}}$ following a specific distribution, i.e., Bernoulli distribution.
This masking strategy is a common but effective strategy in a wide range of previous applications~\cite{you2020graph,thakoor2021bootstrapped}.
We consider this node masking operation as a random data augmentation in graph CL methods~\cite{zhu2021graph,sun2019infograph}.
Then the node feature $\hat{x_i}$ for node $v_i\in \widetilde{\bm{V}}$ in the masked feature mask $\hat{\bm{X}}$ can be defined as follow:
\begin{equation}
    \hat{x}_i=\begin{cases}
 0, &  v_i\in \widetilde{\bm{V}}  \\
 x_i, & v_i\in \bm{V}
\end{cases} 
\label{eq:mae mask feature}
\end{equation}
Further, given an encoder $f_{E}$ and GNN decoder $f_D$ define the embedding code:
\begin{equation}
    \bm{H_1}=f_E(\bm{A},\hat{\bm{X}}) \quad \bm{Z}=f_D(\bm{A},\bm{H_1}),
    \label{eq:mae encoder-decoder}
\end{equation}
where $\bm{H_1} \in \mathbb{R}^{N \times d^h}$ is the hidden embedding of the feature masking view encoded by $f_E$ and then used as the input of GNN decoder $f_d$ to obtain the node representations $\bm{Z}\in \mathbb{R}^{N \times d^{}{}'}$. $d^h$ is the dimension of hidden embedding. Then we calculate the SCE loss:

\begin{equation}
    \label{eq:sce}
    \mathcal{L}_{\mathrm {SCE}}=\frac{1}{|\widetilde{V}|} \sum_{v_i\in\widetilde{V}}(1-\frac{x_i^Th_i}{\|x_i\|\cdot \|z_i\|} )^\gamma ,\gamma >1,
\end{equation}
where $\gamma$ is leveraged to adjust the hyperparameter of loss convergence speed, $h_i$ is the hidden embedding vector of node $v_i$, and $z_i$ is the node representation of node $v_i$.


\subsection{Shared Encoder}

We reveal the intrinsic correlation between CL and Graph MAE and mathematically analyze the feasibility of jointly optimizing them in Section \ref{sec:theory}.
However, how to efficiently transfer global information without complicated structural design is still an unsolved problem.
Therefore, we ask here: 
\begin{quote}
    \textit{how to transfer global information from the contrastive module to the MAE module?}
\end{quote}

An intuitive solution is to train two encoders to encode the CL branch and the MAE branch separately, and then fuse the learned node embeddings.
However, this inevitably faces two issues: \ding{182} training two codes independently cannot realize information transfer between branches. \ding{183} the strategy of fusion embedding will introduce an unnecessary design burden.
Moreover, the accuracy performance depends on the lower bound of either embedding, and low-quality node embeddings will directly affect the performance.

To tackle the problem, we introduce a shared encoder that simultaneously encodes two augmented views to learn local and global information.
The contrastive module leverages the shared encoder to transfer the global information from the entire graph to assist Graph MAE in learning meaningful representations and fine-tuning local network parameters.
We do not need an additional embedding fusion strategy, since the shared encoder can directly yield the unify node embeddings.
Thus, we first use the shared encoder $f_E$ to encode the corrupted graph $\bar{\bm{A}}$ augmented by random 
node dropping into hidden embeddings:
\begin{equation}
    \bm{H_2} = f_E(\bar{\bm{A}},\bm{X}),
    \label{eq:drop encoder}
\end{equation}
where $\bm{H_2} \in \mathbb{R}^{N \times d^h}$.
Then, we utilize two projectors $g_1$, $g_2$ to map $\bm{H_1}$ ,$\bm{H_2}$ into different vector spaces:

\begin{equation}
    \label{eq:projectors}
    \bm{U}=\sigma (g_1(\bm{H_1},\psi_1)),\quad\bm{V}=\sigma (g_2(\bm{H_2},\psi_2)),
\end{equation}
where $\sigma(\cdot)$ is a nonlinear activation function.
The projector consists of a simple two-layer perceptron model with bias $\psi_1,\psi_2$.
We use the classical InfoNCE as the contrastive loss function, defined for each positive sample pair $(u_i,v_i)$ as:
\begin{equation}
    \label{eq:InfoNCE}
    \ell_c=-\log{\frac{\exp (s(u_i^a,v_i^b)/ \tau)}{\sum_{j=1}^{n}[\exp (s(u_i^a,v_j^a))/\tau +\exp(s(u_i^a,v_j^b)/\tau)] } }, 
\end{equation}
where $\tau$ is the temperature parameter and $s(\cdot)$ is the cosine similarity between $u_i$ and $v_i$, 
while $u_i^a \in \bm{U}$ and $u_j^b \in \bm{V}$ denote the projected vector of $v_i$ and $v_j$, respectively.
Since the two projectors are symmetric, the loss for the node pair $(v_i,u_i)$ is defined similarly. The overall objective of the contrastive module is to maximize the average mutual information of all positive sample pairs in the two views. Formally, it is defined as:
\begin{equation}
    \label{eq:overall InfoNCE}
    \mathcal{L}_C=\frac{1}{2N}\sum_{i=1}^{N}[\ell_c (u_i,v_i)+\ell_c(v_i,u_i)].  
\end{equation}


\subsection{Adjacency Matrix Reconstruction}

Graph data exhibits a unique property of complex topology when compared to images and text, as graph contains complex graph topology and node features.
Several studies have focused on adjacency matrix reconstruction~\cite{hu2019strategies,hu2020gpt}.
Since it not only helps models pay more attention to link relationships but also facilitates the encoder to get rid of the constraints brought by fitting extreme feature values~\cite{jin2020self}.
By incorporating the adjacency matrix, the overall objective lightens the weight of feature reconstruction, thereby eliminating the incomplete learned knowledge caused by a single objective.
Different from MaskGAE~\cite{li2022maskgae} reconstructing limited edges, we reconstruct node features and adjacency matrices at the same time, which can help the model capture the global graph structures rather than particular edges or paths.
In this work, we employ the output representation of the decoder to directly reconstruct the adjacency matrix:
\begin{equation}
\label{eq:adj matrix}
 \ell_{MSE}=\frac{1}{N^2}\sum_{i,j}(\hat{\bm{A}}_{ij}-\bm{A}_{ij} )^2, \quad \hat{\bm{A}}=\left \langle \bm{Z}  , \bm{Z}^T  \right \rangle =\bm{Z}\bm{Z}^T,
\end{equation}
where $\hat{\bm{A}}_{ij}$ is the probability of an edge existing between nodes $v_i$ and $v_j$. The MSE function measures the distance between the reconstructed adjacency matrix and the original adjacency matrix, ensuring consistency between the latent embedding and the input graph structure. However, due to the discretization and sparsity of the adjacency matrix, solely using MSE would make the model overfit to zero values.
Therefore, we adopt Cross-Entropy (CE) to determine the existence of an edge between two nodes:
\begin{equation}
    \label{adj matrix: BCE}
    \ell_{BCE}=-\frac{1}{N^2}\sum_{i,j} [\bm{A}_{ij} \cdot \log \hat{\bm{A}}_{ij}+(1-\bm{A}_{ij})\cdot \log(1-\hat{\bm{A}}_{ij} )  ].
\end{equation}

In addition to the above issues, graph data is low-density and the degree distribution follows a power-law distribution, which cannot be adequately estimated by simple MSE and CE.
The coexistence of both low-degree and high-degree nodes further renders Euclidean spaces invalid and hampers the training speed.
To address this problem, we put forth the Relative Distance (RD) loss function:
\begin{equation}
    \ell_{DIST}=-\log\frac{ {\sum_{{(i,j)\in \bm{E}}}}^{} D(z_i,z_j) }{ {\textstyle \sum_{(i,j)\notin \bm{E}}} D(z_i,z_j)},
\end{equation}
where $D(\cdot,\cdot)$ defines the distance between node $v_i$ and node $v_j$, and $z$ is the node representation of the decoder in the MAE module.
The numerator and denominator in the RD loss function correspond to the sum of distances between adjacent nodes and non-adjacent nodes, respectively.
Reconstructing the original degree distribution is an NP-hard problem.
Inspired by CL, we are no longer obsessed with learning the original target distribution, but shift to a proxy task of evaluating node similarity.
The final adjacency matrix reconstruction error is a combination of the three loss functions mentioned above:
\begin{equation}
    \mathcal{L}_{E}=\ell _{MSE}+\ell_{BCE}+\ell_{DIST} .
    \label{eq:edge reconstruction}
\end{equation}

\subsection{Discrimination Loss}

Graph MAE approaches have achieved good classification performance, however, the presence of difficult-to-distinguish dimensions in the features can easily lead to deteriorated training.
In comparison to the sparsity and discretization characteristics of graph data, pixel and text feature vectors possess a higher information density and more significant discrimination between each other.
The low discrimination in graph data is primarily due to node features being vector representations of text, which are compressed representations of keywords obtained through feature extractors.

Previous research has demonstrated a strong correlation between feature discrimination and the performance of Graph MAE~\cite{hou2023graphmae2}.
Graph MAE may mislead the model and facilitate the propagation of harmful substances when reconstructing node features, as numerous information may be related to erroneous evaluations.
In other words, graph MAE attempts to reconstruct the cancerous node features, which results in model collapse and drives the learned representations to be similar.
This observation further inspires us to design a novel loss function to narrow down the gap between Graph MAE and feature discrimination.

To address the above challenges, we introduce a variance-based discrimination loss, which aims to assist the encoder in learning discriminative embeddings and cushion the impact caused by erroneous information to the network.
More importantly, this variance term enforces different node representations within the same embedding matrix are diverse, compensating for the lack of original feature discrimination.    
The regularized standard deviation is defined as follows:
\begin{equation}
    \mathcal{L}_{Var} (h,\epsilon )=\sqrt{\mathrm {Var}(h)+\epsilon},
    \label{eq:variance}
\end{equation}
where $\epsilon$ is a small scalar to prevent numerical instability when reducing to 0 values. $\mathrm {Var}(h)$ is the variance of hidden embeddings (i.e., the output of the shared encoder).
This regularization encourages the encoder to map inputs to a different space within a specific range of variances, thereby preventing the model from collapsing to the same vector. 
 

\begin{algorithm}[t] \caption{The training procedure of GCMAE}
\label{alg:pseudocode}
\begin{algorithmic}[1]
\Require Graph $G=(\bm{A},\bm{X})$, shared encoder $f_E$, decoder $f_D$, projectors $g_1,g_2$, feature masking $\mathcal{T}_1$, node dropping $\mathcal{T}_2$, Epochs $E$
\Ensure Trained GNN encoder $f_E$
\State Initialize the GNN encoder $f_E$
\For {each $epoch \in E$}
\State Generate corrupted views $\mathcal{T}_{1}(G)$, $\mathcal{T}_{2}(G)$
\State $\bm{H_1}=f_E(\mathcal{T}_{1}(G))$, \quad $\bm{H_2}=f_E(\mathcal{T}_{2}(G))$
\State Decoding from hidden embedding $\bm{Z}=f_D(\bm{A},\bm{H_1})$
\State $\bm{U}=g_1(\bm{H_1}), \quad \bm{V}=g_2(\bm{H_2})$ 
\State Calculate loss function $\mathcal{J}=\mathcal{L}_{SCE}+\mathcal{L}_{C}+\mathcal{L}_{E}+\mathcal{L}_{Var}$
\State Update $f_E$ with $\mathcal{J}$
\EndFor
\end{algorithmic}
\end{algorithm}

Algorithm \ref{alg:pseudocode} summarizes the overall training process of \method~ with all the loss terms.

\subsection{Limitations}

Although \method~ enjoys the advantages of both CL and MAE and thus provides strong performance as we will show in Section~\ref{sec:exp}, it still reveals several limitations.
One drawback of \method~ is that its training time may be relatively long because it uses two branches for CL and MAE and learns to reconstruct the entire adjacency matrix. 
As the adjacency matrix contains many edges for large graphs, the time consumption could be high. 
To alleviate this problem, we sample multiple sub-graphs from the original graph for reconstruction. As we will report in Section~\ref{sec:efficiency}, the training time of \method~ is comparable to the baseline methods. 

\section{Experimental Evaluation}\label{sec:exp}

In this part, we extensively evaluate our \method~ along with state-of-the-art baselines on 4 graph tasks, i.e., node classification, link prediction, node clustering, and graph classification. We aim to 
to answer the following research questions:

\begin{itemize}
\item RQ1: How does \method~ compare with the baselines in terms of the \textit{accuracy for the graph tasks}? 
\item RQ2: Does \method~ \textit{achieve its design goals}, i.e., improving MAE with CL?
\item RQ3: How \textit{efficient} it is to train \method~?
\item 
RQ4: How do \textit{our designs and the hyper-parameters} affect the performance of \method~?
\end{itemize}

\begin{table}[t]
    \centering
    \caption{Statistics of the datasets used for node classification, link prediction, and node clustering.}
    \label{tab:node dataset}
\begin{tabular}{c|cccc}
\toprule
\textbf{Dataset} & \textbf{\# Nodes} & \textbf{\# Edges} & \textbf{\# Features} & \textbf{\# Classes} \\
\midrule
Cora     & 2,708    & 10,556   & 1,433       & 7          \\
Citeseer & 3,327    & 9,228    & 3,703       & 6          \\
PubMed   & 19,717   & 88,651   & 500         & 3          \\
Reddit       & 232,965  & 11,606,919 & 602      & 41        \\
\bottomrule
\end{tabular}
\end{table}

\subsection{Experiment Settings}

\stitle{Datasets.}
We conduct the experiments on 10 public graph datasets, which are widely used to evaluate graph self-supervised learning methods~\cite{velickovic2019deep,hou2022graphmae,zhu2020deep}. In particular, the 4 citation networks in Table~\ref{tab:node dataset} are used for node classification, link prediction, and node clustering, and the 6 graphs in Table~\ref{tab:graph classification dataset} are used for graph classification. Note that each dataset in Table~\ref{tab:node dataset} is a single large graph while each dataset in Table~\ref{tab:graph classification dataset} contains many small graphs for graph classification. 
We intentionally choose these graphs for diversity, i.e. they have nodes ranging from thousands to millions, with varying classes.

\stitle{Baselines.}
We compare \method~with 14 state-of-the-art methods on the 4 graph tasks, which span the following 3 categories:
\begin{itemize}
    \item \textbf{Supervised methods} directly train their models with labeled data, and we choose 2 classical GNN models, i.e.,  GCN~\cite{kipf2016semi} and GAT~\cite{velivckovic2017graph}, for node classification.
    
    \item \textbf{Contrastive methods} learn to generate the node embeddings by discriminating positive and negative node pairs. Then, the embeddings serve as input for a separate model (e.g., SVM), which is tuned for the downstream task with labeled data.
    Following GraphMAE~\cite{hou2022graphmae}, we use DGI~\cite{velickovic2019deep}, MVGRL~\cite{hassani2020contrastive}, GRACE~\cite{zhu2020deep}, and CCA-SSG~\cite{zhang2021canonical} for node classification, link prediction, and node clustering. 
    For graph classification, we choose 4 graph-level contrastive methods, i.e., Infograph~\cite{sun2019infograph}, GraphCL~\cite{you2020graph}, JOAO~\cite{you2021graph}, and InfoGCL~\cite{xu2021infogcl}.
    
    \item \textbf{Masked autoencoder (MAE) methods} adopt a ``mask-reconstruct'' structure to learn the node embeddings, and like the contrastive methods, a separate model is tuned for each downstream task. Following SeeGera~\cite{li2023seegera}, we use 4 graph MAE models, i.e.,  GraphMAE~\cite{hou2022graphmae}, SeeGera~\cite{li2023seegera}, S2GAE~\cite{tan2023s2gae}, and MaskGAE~\cite{li2022maskgae} for node classification, link prediction, clustering, and graph classification.
    
    \item \textbf{Deep clustering methods} are specially designed for node clustering and design objectives tailored for clustering to guide model training. 
    We include 3 such methods, i.e., GC-VGE~\cite{guo2022graph}, SCGC~\cite{liu2023simple}, and GCC~\cite{fettal2022efficient}.
\end{itemize}
Note that some baselines may not apply for a task, e.g., the supervised methods only work for node classification, and thus we apply the baselines for the tasks when appropriate. For the contrastive and MAE methods, we use LIBSVM~\cite{chang2011libsvm} to train SVM classifiers for node classification and graph classification following GraphMAE~\cite{hou2022graphmae} and SeeGera~\cite{li2023seegera}. 5-fold cross-validation is used to evaluate performance for the tasks.
For link prediction, we fine-tune the final layer of the model using cross-entropy following MaskGAE~\cite{li2022maskgae}.
For node clustering, we apply  \textit{K-means}~\cite{arthur2006slow} on the node embeddings. We use the Adam optimizer with a weight decay of 0.0001 for our method, and set the initial learning rate as 0.001. We conduct training on one NVIDIA GeForce RTX 4090 GPUs with 24GB memory.

\begin{table}[t]
    \centering
    \caption{Statistics of the datasets used for graph classification.}
    \label{tab:graph classification dataset}
\begin{tabular}{c|ccc}
\toprule
\textbf{Dataset} & \quad \textbf{\# Graphs}  \quad &  \quad \textbf{\# Classes}  \quad &  \quad \textbf{Avg. \# Nodes}  \quad \\
\midrule
IMDB-B   & 1,000     & 2          & 19.8          \\
IMDB-M   & 1,500     & 3          & 13            \\
COLLAB   & 5,000     & 3          & 74.5          \\
MUTAG    & 188       & 2          & 17.9          \\
REDDIT-B & 2,000     & 2          & 429.7         \\
NCI1     & 4,110     & 2          & 29.8          \\
\bottomrule
\end{tabular}
\end{table}

\stitle{Performance metrics.} We are mainly interested in the accuracy of the methods and use well-established accuracy measures for each task~\cite{hou2022graphmae}. In particular, we adopt the Accuracy score (ACC) for node classification, the Area Under the Curve (AUC) and Average Precision (AP) for link prediction, Normalized Mutual Information (NMI) and Adjusted Rand Index (ARI) for node clustering, and Accuracy score (ACC) for graph classification. For all these measures, larger values indicate better performance. For each case (i.e., dataset plus task), we report the average accuracy and standard deviation for each method over 5 runs with different seeds.

\begin{table*}[ht]
\centering
\caption{ Node classification accuracy scores for the experimented methods. For each graph, we mark the most accurate method in \textbf{boldface} and the runner-up with \uline{underline}.}
\label{tab:node classification}
\begin{tabular}{c|c|c|c|c|c}
\toprule
                                                 & Method    & Cora     & Citeseer & PubMed      & Reddit\\
\midrule
\multirow{2}{*}{Supervised}                      & GCN       & 81.48±0.58    & 70.34±0.62     & 79.00±0.50      & 95.30±0.10 \\
                                                 & GAT       & 82.99±0.65 & 72.51±0.71 & 79.02±0.32  & 96.00±0.10 \\
\midrule                                                 
\multirow{4}{*}{Contrastive}                     & DGI       & 82.36±0.62 & 71.82±0.76 & 76.82±0.62  & 94.03±0.10 \\
                                                 & MVGRL     & 83.48±0.53 & 73.27±0.56 & 80.11±0.77  & OOM\\
                                                 & GRACE     & 81.86±0.42 & 71.21±0.53 & 80.62±0.43  & 94.72±0.04\\
                                                 & CCA-SSG   & 84.03±0.47 & 72.99±0.39 & 81.04±0.48  & 95.07±0.02\\
\midrule
\multirow{4}{*}{MAE}                             & GraphMAE  & 85.45±0.40 & 72.48±0.77 & 82.53±0.14  & 96.01±0.08\\
                                                 & SeeGera   & 85.56±0.25 & 72.81±0.13 & 83.01±0.32  &  95.66±0.30\\
                                                 & S2GAE     & 86.15±0.25 & 74.54±0.06 & \uline{86.79±0.22} & 95.27±0.21\\
                                                 & MaskGAE   & \uline{87.31±0.05} & \uline{75.10±0.07} & 86.33±0.26 & \uline{95.17±0.21} \\
\midrule                                                
\multirow{1}{*}{ConMAE}                          & GCMAE & \textbf{88.82±0.11} & \textbf{76.77±0.02} & \textbf{88.51±0.18} & \textbf{97.13±0.17} \\
\bottomrule
\end{tabular}
\end{table*}

\begin{table*}[ht]
\centering
\caption{Link prediction accuracy for the experimented methods. For each graph, we mark the most accurate method in \textbf{boldface} and the runner-up with \uline{underline}. }
\label{tab:link prediction}
\setlength{\tabcolsep}{3pt}
\begin{tabular}{c|c|cc|cc|cc|cc}
\toprule
& \multirow{2}{*}{Method}            & \multicolumn{2}{c}{Cora}    & \multicolumn{2}{c}{Citeseer} & \multicolumn{2}{c}{PubMed} & \multicolumn{2}{c}{Reddit} \\
\cmidrule{3-10}
           & & AUC          & AP           & AUC           & AP           & AUC          & AP   & AUC          & AP         \\           
\midrule
\multirow{4}{*}{Contrastive} &DGI         & 93.88±1.00 & 93.60±1.14 & 95.98±0.72  & 96.18±0.68 & 96.30±0.20 & 95.65±0.26 & 97.05±0.42 & 96.74±0.16\\
&MVGRL       & 93.33±0.68 & 92.95±0.82 & 88.66±5.27  & 89.37±4.55 & 95.89±0.22 & 95.53±0.30 &OOM &OOM\\
&GRACE       & 93.46±0.71 & 92.74±0.48 & 92.07±0.51 & 90.32±0.57   & 96.11±0.13 & 95.37±0.25 & 95.82±0.24 & 95.74±0.46 \\
&CCA-SSG     & 93.88±0.95 & 93.74±1.15 & 94.69±0.95  & 95.06±0.91 & 96.63±0.15 & 95.97±0.23 & 97.74±0.20 & 97.58±0.12 \\
\midrule
\multirow{4}{*}{MAE} &GraphMAE    & 90.70±0.01 & 89.52±0.01 & 70.55±0.05  & 74.50±0.04 & 69.12±0.01 & 87.92±0.01 & 96.85±0.24 & 96.77±0.35\\
&SeeGera     & 95.50±0.71 & 95.92±0.68 & 97.04±0.47  & 97.33±0.46 & 97.87±0.20 & 97.88±0.21 & - & -\\
&S2GAE       & 95.05±0.76 & 95.01±0.62 & 94.85±0.49  & 94.84±0.23 & 98.45±0.03 & 98.22±0.05 & 97.02±0.31 & 97.10±0.27\\
&MaskGAE     & \uline{96.66±0.17}   & \uline{96.29±0.23}   & \uline{98.00±0.23}    & \uline{98.25±0.16}   & \uline{99.06±0.05}   & \textbf{98.99±0.06} &\uline{97.75±0.20} & \uline{97.67±0.14}  \\
\midrule
\multirow{1}{*}{ConMAE} &GCMAE & \textbf{98.00±0.03}   & \textbf{97.74±0.37}   & \textbf{99.48±0.18}    & \textbf{99.46±0.23}  & \textbf{99.14±0.27} & \uline{98.82±0.13} & \textbf{98.87±0.18}   &\textbf{98.62±0.26}\\
\bottomrule
\end{tabular}

\end{table*}

\begin{table*}[ht]
\centering
\caption{Node clustering accuracy for the experimented methods. For each graph, we mark the most accurate method in \textbf{boldface} and the runner-up with \uline{underline}.}
\label{tab:node clustering}
\begin{tabular}{c|c|cc|cc|cc|cc}
\toprule
                        & \multirow{1}{*}{Method} & \multicolumn{2}{c}{Cora} & \multicolumn{2}{c}{Citeseer} & \multicolumn{2}{c}{PubMed} & \multicolumn{2}{c}{Reddit}  \\
\cmidrule{3-10}
                        & & NMI         & ARI        & NMI           & ARI          & NMI          & ARI          & NMI          & ARI       \\
                        \midrule
\multirow{4}{*}{Contrastive}&DGI                     & 52.75±0.94  & 47.78±0.65 & 40.43±0.81    & 41.84±0.62   & 30.03±0.50   & 29.78±0.28 &66.87±0.33 & 64.27±0.25 \\
&MVGRL                   & 54.21±0.25  & 49.04±0.67 & 43.26±0.48    & 42.73±0.93   & 30.75±0.54   & 30.42±0.45 & OOM & OOM  \\
&GRACE                   & 54.59±0.32 & 48.31±0.63 & 43.02±0.43 & 42.32±0.81   & 31.11±0.48 & 30.37±0.51 & 65.24±0.24  & 63.60±0.38\\
&CCA-SSG                 & 56.38±0.62  & 50.62±0.90 & 43.98±0.94    & 42.79±0.77   & 32.06±0.40   & 31.15±0.85 & 68.09±0.24 & 67.73±0.37  \\
\midrule
\multirow{3}{*}{MAE}&GraphMAE                & 58.33±0.78  & 51.64±0.41 & 45.17±1.12    & 44.73±0.55   & 32.52±0.53   & 31.48±0.39 &65.82±0.13 &  64.43±0.15 \\
&S2GAE                   & 56.25±0.43  & 50.21±0.44 & 44.82±0.56    & 44.51±0.94   & 31.48±0.35   & 30.86±0.60 &66.00±0.40 &65.95±0.19  \\
&MaskGAE                 & 59.09±0.26  & 52.19±0.51 & \uline{45.46±0.77}    & 45.68±0.42   & \uline{33.91±0.35}   & \uline{32.64±0.68} & \uline{68.24±0.13} & \uline{67.80±0.04} \\ 
\midrule
\multirow{3}{*}{Clustering}&GC-VGE                  & 53.57±0.30       & 48.15±0.45      & 40.91±0.56         & 41.51±0.32        & 29.71±0.53        & 29.76±0.66 &53.58±0.15 &51.91±0.18      \\
&SCGC                    & 56.10±0.72  & 51.79±1.59 & 45.25±0.45    & \uline{46.29±1.13}   &       -       &       -    & - & -  \\
&GCC                     & \uline{59.17±0.28}  & \uline{52.57±0.41} & 45.13±0.68    & 45.05±0.93   & 32.30±0.48   & 31.23±0.44 & 62.35±0.24 & 60.40±0.16 \\
\midrule
\multirow{1}{*}{ConMAE}&GCMAE             & \textbf{59.31±0.12}  & \textbf{52.98±0.21} & \textbf{45.84±0.58}    & \textbf{46.54±0.89}   & \textbf{34.98±0.85}   & \textbf{33.76±0.61}  & \textbf{69.79±0.11} & \textbf{69.28±0.19}\\
\bottomrule
\end{tabular}
\end{table*}

\begin{table*}[ht]

\centering
\caption{Graph classification accuracy for the experimented methods. For each dataset, we mark the most accurate method in \textbf{boldface} and the runner-up with \uline{underline}.}
\label{tab:graph classification}
\begin{tabular}{c|c|ccccccc}
\toprule
&\multirow{1}{*}{Method}            & IMDB-B     & IMDB-M       & COLLAB     & MUTAG      & REDDIT-B   & NCI1       \\
\midrule
\multirow{4}{*}{Contrastive}&Infograph   & 73.03±0.87 & 49.69±0.53  & 70.65±1.13 & 89.01±1.13 & 82.50±1.42 & 76.20±1.06 \\
&GraphCL     & 71.14±0.44 & 48.58±0.67   & 71.36±1.15 & 86.80±1.34 & \uline{89.53±0.84} & 77.87±0.41 \\
&JOAO        & 70.21±3.08 & 49.20±0.77  & 69.50±0.36 & 87.35±1.02 & 85.29±1.35 & 78.07±0.47 \\
&MVGRL       & 74.20±0.70 & 51.20±0.50      &      OOM     & 89.70±1.10 & 84.50±0.60 &        OOM   \\
&InfoGCL     & 75.10±0.90 & 51.40±0.80     & 80.00±1.30 & \uline{91.20±1.30} &    OOM      & 80.20±0.60 \\
\midrule
\multirow{2}{*}{MAE}&GraphMAE    & 75.52±0.66 & 51.63±0.52  & 80.32±0.46 & 88.19±1.26 & 88.01±0.19 & 80.40±0.30 \\
&S2GAE       & \uline{75.76±0.62} & \uline{51.79±0.36}  & \uline{81.02±0.53} & 88.26±0.76 & 87.83±0.27 & \uline{80.80+0.24} \\
\midrule
\multirow{1}{*}{ConMAE}&GCMAE & \textbf{75.78±0.23} & \textbf{52.49±0.45}  & \textbf{81.32±0.32} & \textbf{91.28±0.55} & \textbf{91.75±0.22} & \textbf{81.42±0.30} \\
\bottomrule
\end{tabular}
\end{table*}

\begin{figure}[!t]
    \centering    \includegraphics[width=\linewidth,scale=1.00]{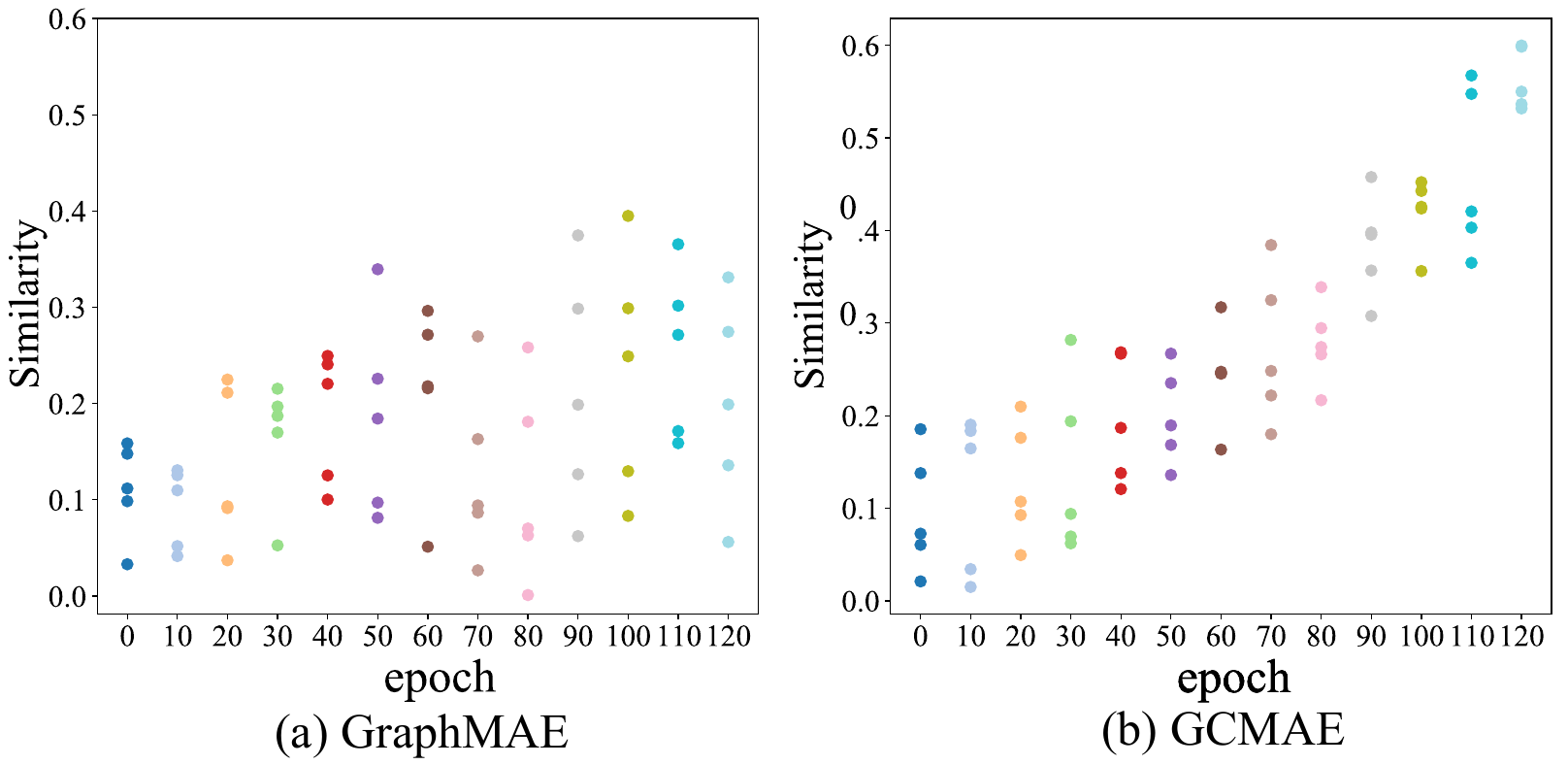}
    \caption{The embedding similarity between a node and its 5-hop neighbors w.r.t. the number of training epochs.}
    \label{fig:global_info}
\end{figure}

\subsection{Accuracy for the Graph Tasks (RQ1)}

\stitle{Node classification.}
\autoref{tab:node classification} reports the accuracy scores of our \method~and the baselines for node classification. We observe that \method~is the most accurate method on all datasets, and compared with the best baseline, the accuracy improvements of \method~ are 1.7\% on Cora, 2.2\% on Citeseer, 2.0\% on PubMed, and 2.1\% on Reddit. Considering the baselines, the supervised methods (i.e., GCN and GAT) perform the worst because they can only utilize label information while the other methods use self-supervised learning to introduce more supervision signals. The graph MAE methods generally perform better than the contrastive methods because node classification relies on the local information of each node (i.e., a local task), and graph MAE is better at capturing local information by learning to reconstruct individual node features and masked edges. Similar pattern is also observed in the results of other local tasks, i.e., link prediction and node clustering. 
The fact that \method~outperforms both contrastive and MAE methods suggests that our model designs allow us to enjoy the benefits of both paradigms.

\stitle{Link prediction.} Following SeeGera~\cite{li2023seegera}, S2GAE~\cite{tan2023s2gae},  we do not report the results of supervised methods for link prediction. 
Since DGI~\cite{velickovic2019deep}, MVGRL~\cite{hassani2020contrastive}, and GraphMAE~\cite{hou2022graphmae} do not report relevant experimental results, we train a network for each of them and output prediction results.

The results of the link prediction are shown in \autoref{tab:link prediction}.
\method~ achieves the best prediction results (except AP on PubMEd), with an average improvement of 1.1\% on AUC and 0.8\% when compared to the runner-up MaskGAE.
The performance of \method~ exceeds all contrastive methods by a large margin, with an average increase of 5.9\% in AUC and 5.6\% in AP.
Compared to contrastive methods, we use adjacency matrix reconstruction as part of the total objective, forcing the model to pay more attention to graph structures.
GraphMAE~\cite{hou2022graphmae} performs poorly on link prediction, which indicates that only reconstructing node features can lead to performance degradation on link-level tasks.
In contrast, MaskMAE~\cite{li2022maskgae} takes edges as reconstruction objectives, which is consistent with downstream tasks, and unsurprisingly becomes the strongest method among all baselines.
Based on the observation, the adjacency matrix reconstruction brings more performance improvement to link prediction than edge reconstruction, because the model can capture more meaningful global structures of the graph.

\begin{table}[!t]
    \centering
    \caption{\textbf{Node classification accuracy when using different designs for the encoder of \method~.}}
\label{tab:shared encoder}
\scalebox{1.0}{
\begin{tabular}{c|ccc}
\toprule
               & \quad Cora \quad     & \quad Citeseer \quad & \quad PubMed  \quad    \\
\midrule
MAE Encoder    & 84.14  & 73.17    & 81.83 \\
Con. Encoder   & 68.46 & 60.46    & 57.61 \\
Fusion Encoder         & 85.61 & 71.71     & 78.63 \\
Shared Encoder & \textbf{88.82} & \textbf{76.77}    & \textbf{88.51} \\
\bottomrule
\end{tabular}
}
\end{table}

\stitle{Node clustering.}
As shown in \autoref{tab:node clustering}, \method~ achieves the best results among all baselines on the node clustering task.
Especially on Citeseer, \method~ improves by 2.2\% on Reddit w.r.t. ARI when compared to the runner-up approach MaskGAE~\cite{li2022maskgae}.
Compared with the MAE method, the improvements range from 0.2\% to 3.2\% regarding NMI and from 0.5\% to 3.4\% regarding ARI.
Since \method~ can learn the feature and structure difference between nodes of different clusters from global information, which helps the model clarify the boundaries of clusters.
We can observe that the performance gap between the contrastive method and the MAE method on node clustering is not large, unlike node classification and link prediction.
This is because the goal of node clustering is to divide the data set into different clusters, in other words, maximizing the similarity of intra-cluster nodes and expanding the difference of inter-cluster nodes. This goal is similar to the intrinsic mechanism of CL.
Moreover, we choose 3 deep node clustering methods as the baseline, and \method~ can still achieve an average improvement of 10.5\% in NMI and 11.4\% in ARI.
This means that we can obtain high-quality node embeddings for the node clustering task without deliberately tailoring a clustering loss to guide the training process.

\stitle{Graph classification.}
SeeGera~\cite{li2023seegera} and MaskGAE~\cite{li2022maskgae} are not chosen as baselines due to the unavailable source code.
\autoref{tab:graph classification} reports all the experimental results for graph classification.
We can see that the contrastive methods and the graph MAE methods have comparable performance in graph classification, each achieving three runner-up results. 
In contrast, \method~ achieves the highest accuracy on all datasets when compared to all baselines, the accuracy of our method is improved by an average of 1.3\%.
Our method benefits from both MAE and CL, and therefore can effectively distinguish the differences between graph-level embeddings by comparing multiple corrupted views.
Compared to GraphMAE~\cite{hou2022graphmae}, by enhancing the feature discrimination through discrimination loss, our proposed method is able to learn meaningful representations even from features with limited information content, such as one-hot vectors based on labels or degrees.
Overall, \method~ does not only excel on node-level (i.e., node classification and node clustering) and link-level (i.e., link prediction) tasks, and it can generalize well to graph-level downstream tasks.
The results clearly demonstrate the effectiveness of the proposed \method~ framework and validate our claims.

\subsection{Anatomy of the Design Goals (RQ2)}

\stitle{\method~ learns global information.}
To verify whether CL can help graph MAE to obtain global information, we visualize the similarity between nodes in GraphMAE and our \method~, respectively.
Specifically, we randomly select distant nodes 5 hops away from the target node and then calculate the similarity between them. 
We can observe that as the training epoch increases, the similarity of target nodes to distant nodes remains at a low level, which means that MAE is prone to learn node embeddings from local information, according to the result shown in \autoref{fig:global_info} (a).
When we combine the CL and graph MAE, \method~ can gradually improve the similarity between target nodes to distant nodes.
In other words, CL can make up for the shortcoming of GNN layers that are too shallow and potentially help graph MAE surpass the constraints of local receptive field and acquire global information, letting graph nodes gain useful knowledge from nodes or edges that are out of GNN’s aggregation scope.

Note that as the number of training epochs increases, the similarity between the target node and the distant node does not continue to increase. As shown in \autoref{fig:global_info}, after a certain number of epochs (i.e., epoch=90), the similarity tends to be stable in (0.4-0.6). Therefore, the model will not face the over-smoothing issue.

\stitle{The shared encoder effectively transfers global information.}
We study the impact of shared encoder and unshared encoder on our model to evaluate whether the shared encoder can pass global information.
Therefore, we conduct the node classification on Cora, Citeseer, and PubMed with the different types of encoder, \autoref{tab:shared encoder} presents the accuracy results.
We can find that ``MAE Encoder'' outperforms the other two independently parameterized encoders by a significant margin, but does not surpass the shared-parameter encoder.
Kindly note that only using ``MAE Encoder'' means our method degenerates to GraphMAE~\cite{hou2022graphmae}.
The ``Con. Encoder'' does not perform as well as expected, which may be due to the excessive corruption of the input graph caused by a high mask ratio, leading to the failure of the contrastive encoder.
``Fusion Encoder'' represents the average sum of the embeddings generated by the ``MAE Encoder'' and ``Con. Encoder''.
Naturally, ``Fusion Encoder'' may suffer from collapsed contrastive encoder, which results in suboptimal results.
This is particularly important that the ``Shared Encoder'' achieves the best classification performance, which means CL can convey global information to the MAE module through the shared-parameter encoder, aiding MAE in perceiving long-range node semantics.

\subsection{The Training Efficiency of \method (RQ3)}
\label{sec:efficiency}

In order to study whether unifying CL and graph MAE will increase the training time consumption, we conduct node classification on 4 datasets: Cora, Citeseer, PubMed, and Reddit, and report the total time consumption.
We choose CCA~\cite{zhang2021canonical} with the best performance on node classification among the contrastive methods, backbone method GraphMAE~\cite{hou2022graphmae} and MaskGAE~\cite{li2022maskgae} with the best performance among MAE methods as comparison methods.
\autoref{tab:time cost} shows the time consumption of all methods under the parameter setting with the highest node classification accuracy, that is, the sum of pre-training time and the fine-tuning time for downstream tasks.

\begin{table}[t]
\centering
\caption{End-to-end training time of representative methods. ``s'' means seconds and ``h'' means hours.}
\label{tab:time cost}
\begin{tabular}{c|cccc}

\toprule
  Method          & Cora    & Citeseer & PubMed   & Reddit \\
\midrule            
CCA-SSG     & \textbf{2.2(s) }  & \textbf{1.9(s)}    & \textbf{4.6(s)}     & \textbf{0.8(h)}   \\
GraphMAE    & 152.8(s) & 93.1(s)   & 1270.1(s) & 18.2(h)   \\
MaskGAE     & 26.3(s)  & 40.5(s)   & 52.7(s)   & 2.3(h)   \\
\method~ & 28.6(s)  & 55.3(s)  & 508.9(s)  & 2.5(h)   \\
\bottomrule

\end{tabular}
\end{table}

We can observe that CCA-SSG has the least time consumption, which is due to the use of \textit{canonical correlation analysis} to optimize the calculation between the embeddings of the two views, which greatly reduces the time consumption caused by large matrix operations.
Our \method~ is on average 2$\times$ faster than GraphMAE.
This is because GraphMAE uses GAT~\cite{velivckovic2017graph} as an encoder, and GAT needs to take the entire adjacency matrix as input when encoding the input graph.
Even if we try to reduce the dimension of the hidden embeddings, it still introduces unacceptable time consumption when encountering large-scale graphs (e.g., Reddit with millions of nodes).
Unlike GraphMAE, the overall time consumption of our method is similar to that of MaskGAE, because we both use GraphSAGE~\cite{hamilton2017inductive} to encode node embeddings, which can sample multiple subgraphs from a large-scale graph for mini-batch training without inputting the entire graph into the network.
However, our method is still slower than MaskGAE, because we reconstruct the entire adjacency matrix instead of only reconstructing partial edges like MaskGAE.
Overall, the efficiency performance of \method is comparable to prior works,
and there is not a significant increase in time consumption due to the combination of graph MAE and CL.

\begin{table}[t]
\centering
\caption{Node classification accuracy when removing the key components of \method.}
\label{tab:ablation on framework }
\scalebox{1.0}{
\begin{tabular}{c|ccc}
\toprule
              & \quad Cora \quad     & \quad Citeseer \quad & \quad PubMed  \quad
  \\
\midrule
GCMAE    & \textbf{88.8} & \textbf{76.7} & \textbf{88.5} \\
w/o Contrast.        & 87.3 & 75.7 & 87.4 \\
w/o Stru. Rec. & 86.0 & 73.5 & 86.7 \\
w/o Disc. Loss  & 87.0 & 74.1 & 86.9 \\
GraphMAE       & 85.5 & 72.5 & 82.5 \\
\bottomrule
\end{tabular}
}
\end{table}
\subsection{Ablation Study and Parameters (RQ4)} 

In this part, we conduct an ablation study for the designs of \method~ and explore the influence of the parameters. We experiment with the task of node classification and note that the observations are similar for the other graph tasks.

\stitle{The components of \method.}
To study the effectiveness of each component, we compare our complete framework with GraphMAE~\cite{hou2022graphmae} and three variants: ``w/o Con.'', ``w/o Stru. Rec.'' and ``w/o Disc.''.
The results are presented in \autoref{tab:ablation on framework }, where ``w/o Con.'' means the removal of the CL loss, ``w/o Stru. Rec.'' is our method without adjacency matrix reconstruction, and ``w/o Disc.'' is our method without feature discrimination loss.
All results correspond to the accuracy values (\%) of node classification tasks on 3 benchmark datasets: Cora, Citeseer, and PubMed.
We can find that the removal of any of these components leads to a decrease in performance.
In particular, the exclusion of adjacency matrix reconstruction has the most severe impact on the model, resulting in a decrease of 3.2\% in Cora, 4.3\% in Citeseer, and 2.4\% in PubMed.
Interestingly, even if we totally remove the contrastive module, our method still outperforms the GraphMAE~\cite{hou2022graphmae}. This is because the adjacency matrix reconstruction provides rich graph structure information and the discrimination loss improves the feature discrimination among nodes.
In other words, adjacency matrix reconstruction and discrimination loss play an extremely crucial role in our framework.
In summary, all the above three components contribute significantly to the final results.

\begin{figure}[t]
    \centering
    \includegraphics[width=\linewidth,scale=0.9]{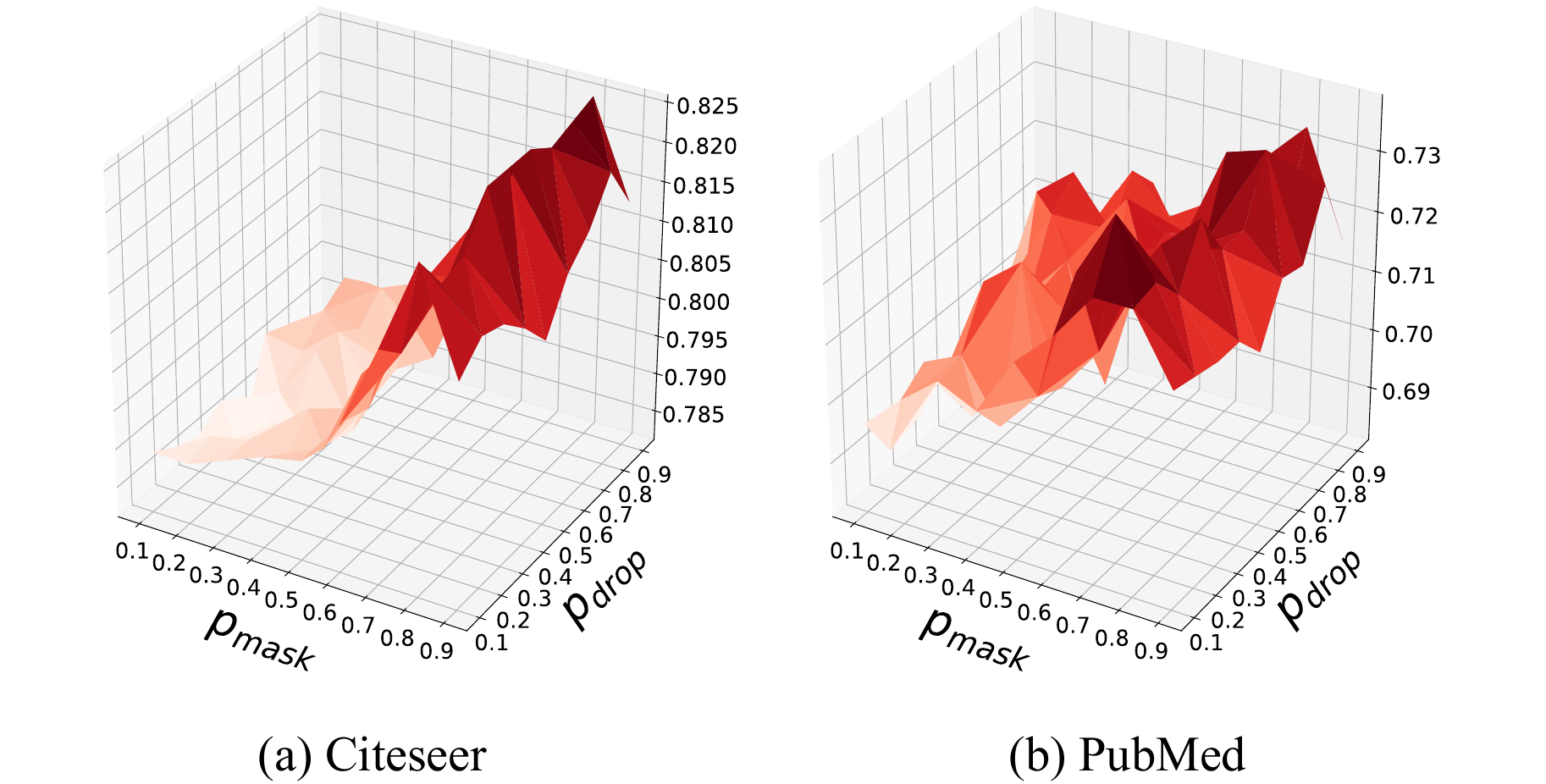}
    \caption{The influence of edge mask rate $p_{mask}$ and node drop node rate $p_{drop}$ on the performance of node classification.}
    \label{fig:F1}
\end{figure}

\stitle{Effect of the hyper-parameters.}
To study the influence of different hyper-parameter settings on the model, we conduct sensitivity experiments on mask rate and drop rate in GCMAE.
We report the performance in \autoref{fig:F1}, where the $x$-axis, $y$-axis, and $z$-axis represent the feature mask rate $p_{mask}$, the drop node rate $p_{drop}$, and the F1-Score value, respectively.
We can find that the variation trends on all datasets are consistent.
When $p_{mask}$ is large (0.5-0.8), the model performance remains within a satisfactory range, which can also be observed in previous graph MAE models.
A higher mask rate means lower redundancy, in which can help the encoder recover missing node features from the few neighboring nodes.
When $p_{mask}$ is fixed, increasing $p_{drop}$ can improve the classification accuracy.
Overall, compared to $p_{drop}$, $p_{mask}$ plays a decisive role in model performance, the variation of $p_{mask}$ directly affects the experiment performance, while changes in $p_{drop}$ do not cause significant fluctuations in the final results.

\begin{figure}[t]
    \centering
    \includegraphics[width=\linewidth,scale=0.9]{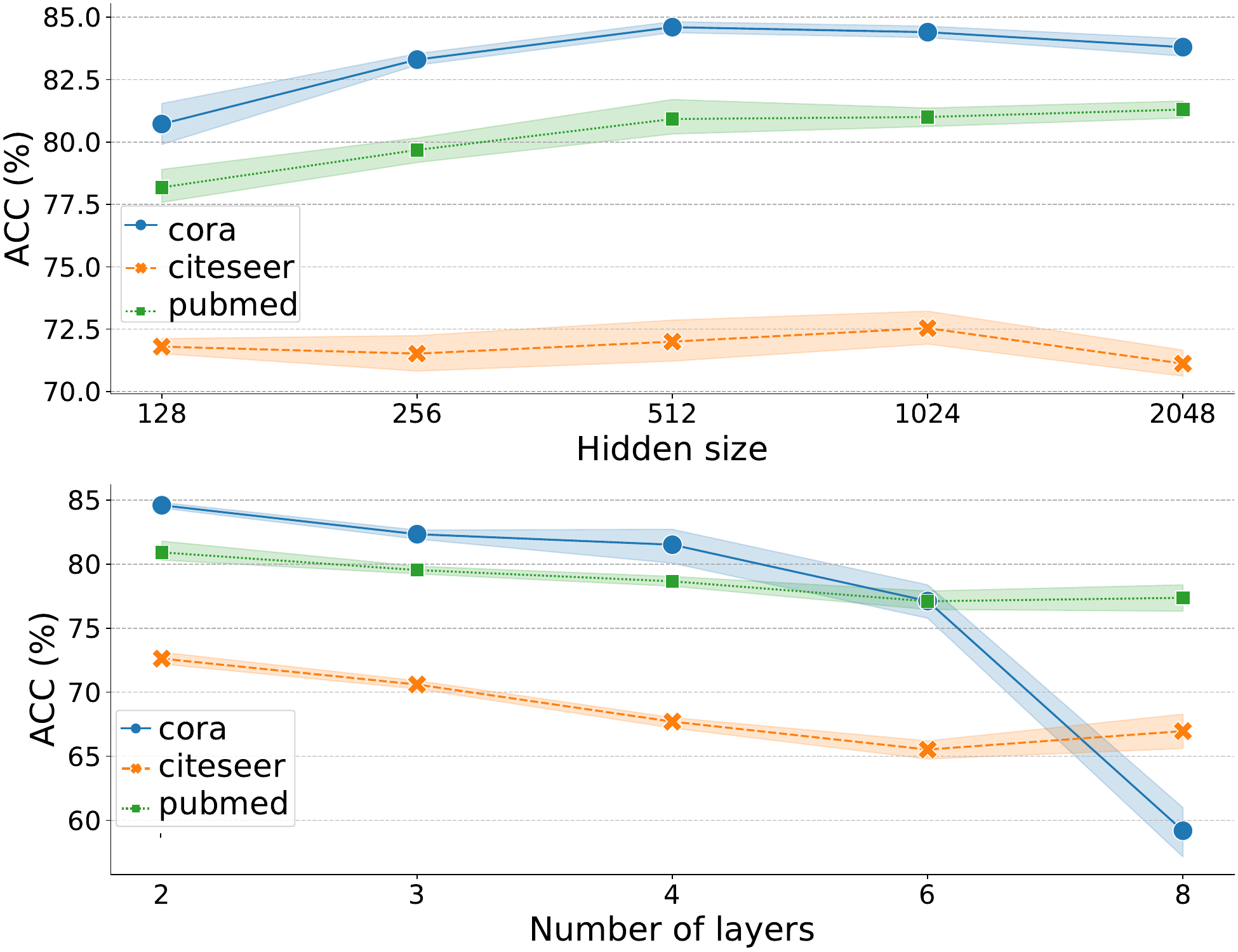}
    \caption{The influence of the hidden layer dimension and the number of layers in the GNN models on the accuracy of our \method~ for node classification.}
    \label{fig:size_layer}
\end{figure}

The impact of network width and depth on performance has attracted significant attention in the CL methods~\cite{zhu2020deep,mo2022simple}.
Therefore, we investigate the effects of selecting multiple network scale parameters on our model.
As shown in \autoref{fig:size_layer}, increasing the network width results in a performance improvement of approximately 5.9\% in Cora, 3.1\% in Citesser, and 4.0\% in PubMed, as long as the hidden size does not exceed 1024 (except for PubMed where the limit is 2048).
In contrast, a smaller hidden size leads to a sharp drop in performance.
GCMAE achieves the best performance on most benchmarks with a 512-dimensional embedding.
This means a moderate hidden size is crucial for the model to learn informative and compact node embeddings for downstream tasks.

Meanwhile, we can observe that network depth has a relatively smaller impact on performance compared to network width.
\autoref{fig:size_layer} shows that when the network has 2 layers, the accuracy is highest across all datasets.
As the depth increases, the performance gradually decreases, with a decline of 30.0\% in Cora, 7.8\% in Citeseer, and 3.5\% in PubMed when the depth reaches 8.
Surprisingly, GNNs like GIN and GAT achieve good results with shallow depths, contrary to empirical results in the field of computer vision~\cite{he2022masked}.
This may be due to the fact that deep GNNs are challenging to optimize. 
As the depth increases, the learned representations tend to become homogeneous, thus degrading the performance.

\section{Related Work}\label{sec:related}

In this part, we review representative works on contrastive and generative methods for graph self-supervised learning (GSSL).

\subsection{Contrastive Methods for Graph SSL}
Inspired by the success of CL in 
computer vision and natural language processing ~\cite{chen2020simple,gao2021simcse,he2020momentum}, many works develop constrative methods for graph learning~\cite{velickovic2019deep,hassani2020contrastive,zhu2020deep,xia2021progcl,li2023homogcl}.
These methods generally produce multiple corrupted views of the graph via data augmentation and maximize the similarity between these views.
For instance, GraphCL~\cite{you2020graph} adopts four types of graph augmentations (i.e., node dropping, edge perturbation, attribute masking, and subgraph sampling) to incorporate different priors and learns to predict whether two graphs are generated from the same graph.
To improve GraphCL, GCA~\cite{zhu2021graph} determines the importance of each node via its centrality and decides whether to mask a node according to its importance. The node with higher centrality is less likely to be masked. BGRL~\cite{thakoor2021bootstrapped} proposes to conduct CL without negative samples and thus reduces model training time.
DGI~\cite{velickovic2019deep} conducts CL using patches of a graph and uses a read-out function to compute graph-level embedding from the node embeddings. GRACE~\cite{zhu2020deep} corrupts both the graph topology and the node features such that contrast learning can capture more information.
MVGRL~\cite{hassani2020contrastive} observes that simply increasing the number of views does not improve performance and proposes to maximize the mutual information among the node and graph representations in different views.
CCA-SSG~\cite{zhang2021canonical} leverages canonical correlation analysis to speedup the computation of contrastive loss among multiple augmented views and reduce model training time. We observe that the contrastive methods are good at capturing global information of the graph but poor in learning the local information for particular edges and nodes. Thus, by augmenting MAE with CL, our \method~outperforms the existing GSSL methods.

\subsection{Generative SSL Method}
Different from contrastive methods, generative methods aim to reconstruct input data from hidden embeddings via a decoder and then minimize the distance between the input graph and the reconstructed graph.

\textbf{Graph Autoencoder} (GAE) is a classical self-supervised learning method, which encodes the graph structure information into a low-dimensional latent space and then reconstructs the adjacency matrix from hidden embeddings \cite{wang2017mgae,salehi2019graph,cao2016deep}.
For example, DNGR~\cite{cao2016deep} uses a stacked denoising autoencoder~\cite{vincent2008extracting} to encode and decode the PPMI matrix via multi-layer perceptrons.
However, DNGR ignores the feature information when encoding node embeddings.
Therefore, GAE~\cite{kipf2016variational} utilizes GCN~\cite{kipf2016semi} to encode node structural information and node feature information at the same time and then uses a dot-product operation for reconstructing the input graph.
Variational GAE (VGAE)~\cite{kipf2016variational} learns the distribution of data, where KL divergence is used to measure the distance between the empirical distribution and  the prior distribution.
In order to further narrow the gap between the above two distributions, ARVGA~\cite{pan2018adversarially} employs the training scheme of a generative adversarial network~\cite{goodfellow2014generative} to address the approximate problem.
RGVAE~\cite{ma2018constrained} further imposes validity constraints on a graph variational autoencoder to regularize the output distribution of the decoder.
Unlike the previous asymmetric decoder structure, GALA~\cite{park2019symmetric} builds a fully symmetric decoder, which facilitates the proposed autoencoder architecture to make use of the graph structure.
Later studies focused on leveraging feature reconstruction or additional auxiliary information~\cite{pan2018adversarially,salehi2019graph}.
Unfortunately, most of them mainly perform well on a single task such as node classification or link prediction, since they are limited by a sufficient reconstruction objective. 
However, our \method~ surpasses these GAE methods on various downstream tasks due to both reconstructing node features and edge.

\textbf{Masked Autoencoders} learn graph representations by masking certain nodes or edges and then reconstructing the masked tokens~\cite{hu2020gpt,zhang2021graph,hou2023graphmae2}.
This strategy allows the graph to use its own structure and feature information in a self-supervised manner without expensive label annotations.
Recently, GraphMAE~\cite{hou2022graphmae} enforces the model to reconstruct the original graph from redundant node features by masking node features and applying a re-masking strategy before a GNN decoder.
Instead of masking node features, MaskGAE~\cite{li2022maskgae} selects edges as the masked token and then reconstructs the graph edges or random masked path accordingly.
MaskGAE achieves superior performance in link prediction tasks compared to other graph MAE methods, but its performance in classification tasks is not as satisfactory as feature-based MAE because it does not reconstruct the node features.
S2GAE~\cite{tan2023s2gae} proposes a cross-correlation decoder to explicitly capture the similarity of the relationship between two connected nodes at different granularities.
SeeGera~\cite{li2023seegera} is a hierarchical variational framework that jointly embeds nodes and features in the encoder and reconstructs links and features in the decoder, where an additional structure/feature masking layer is added to improve the generalization ability of the model.
Based on the above observations, these graph MAE methods all suffer from inaccessible global information, resulting in sub-optimal performance.
However, \method~ surpasses these graph MAE methods, as unifying CL and graph MAE enjoys the benefits of both paradigms and yields more high-quality node embeddings.

\section{CONCLUSION}
In this paper, we observed that the two main paradigms for graph self-supervised learning, i.e., masked autoencoder and contrastive learning, have their own limitations but complement each other. Thus, we proposed the \method~framework to jointly utilize MAE and contrastive learning for enhanced performance. \method~ comes with tailored model designs including a shared encoder for information exchange, discrimination loss to tackle feature smoothing, and adjacency matrix reconstruction to learn global information of the graph. We conducted extensive experiments to evaluate \method~on various graph tasks. The results show that \method~ outperforms state-of-the-art GSSL methods by a large margin and is general across graph tasks.       


\clearpage
\normalem
\bibliographystyle{ACM-Reference-Format}
\bibliography{ref}

\end{document}